# On Affinity Measures for Artificial Immune System Movie Recommenders

**Uwe Aickelin and Qi Chen,**
School of Computer Science and IT,
University of Nottingham, NG8 1BB, UK
{uxa, qxc}@cs.nott.ac.uk,

**Abstract**. *We combine Artificial Immune Systems (AIS) technology with Collaborative Filtering (CF) and use it to build a movie recommendation system. We already know that Artificial Immune Systems work well as movie recommenders from previous work by Cayzer and Aickelin ([3], [4], [5]). Here our aim is to investigate the effect of different affinity measure algorithms for the AIS. Two different affinity measures, Kendall's Tau and Weighted Kappa, are used to calculate the correlation coefficients for the movie recommender. We compare the results with those published previously and show that that Weighted Kappa is more suitable than others for movie problems. We also show that AIS are generally robust movie recommenders and that, as long as a suitable affinity measure is chosen, results are good.*

## 1 Introduction

Cayzer and Aickelin have previously shown that Artificial Immune System (AIS) perform well as movie recommendation tools([3], [4],[5]). Their research also provided some insight into how certain parameters, such as overlap or the neighbourhood size, influence the AIS results. In this paper we intend to extend their work by focusing on a yet uncovered issue, the choice of the affinity measure. We will show how the choice of affinity measure will influence the AIS results, and try to answer the more general question if there is such a thing as a 'best' affinity measure for the movie recommendation problem.

We use Collaborative Filtering (CF) technology [2] and an AIS [7] to create movie recommendations in our project. CF works by offering the target user recommendations based on other users who have similar preferences [2]. It offers the user recommendations even though specific information regarding the movie is unknown. A group of users, known as a neighbourhood, who have similar preferences to the target user, determine the recommendations. In this project, we use an AIS to choose the members of the neighbourhood.

Two affinity measure algorithms, Weighted Kappa and Kendall's Tau, are adopted respectively to calculate the correlation coefficients between the users. We compare the results obtained from these systems to those obtained previously [5] and achieve an impressive improvement choosing the most suitable measure. The paper concludes with our assessment of why certain methods work better than others and a general recommendation as to which affinity measure is the most suitable.

## 2 AIS and Immune Network Models

AIS are distributed adaptive systems for problem solving using models and principles derived from the Human Immune System [7]. The Immune System is the defence system of our body, which can produce and secrete antibodies used to protect us against infection through an antigen recognition process ([8], [9], [13]). Many different AIS algorithm models have been built, including bone marrow models, thymus models, clonal selection algorithms and immune network models [7].

In this project, we will use the immune network model suggested by Farmer et al [8] and modified and extended by Cayzer and Aickelin ([3], [4], [5]) and by Morrison and Aickelin [11]. In this model, the interaction between antibodies and antibodies, and the interaction between the antibodies and antigens control the system, and maintain a distributed adaptive system with diversity.

Farmer's Immune Network Model is based on matching, diversity, and distributed control ([5], [10]). Matching is the binding between the antibodies and antigens. Diversity has the objective of achieving optimal antigen space coverage and is based on the premise that antibodies can match other antibodies as well as antigens. Distributed control means that there is no central control in the Immune Network Model, as the model is controlled by local interactions between antibodies and antigens. The users in the database are viewed as candidate antibodies, and the target user who uses the movie recommendation system is the antigen. Using statistical correlation methods we will calculate the correlations (also termed affinities) between the antigen-antibodies and antibodies-antibodies [5].

Briefly, the AIS model used in this project works as follows:
(1) Choose a fixed number (n=100) of antibodies as the initial AIS.
(2) Calculate the concentration of the antibodies in the AIS based on their affinity to the antigen and all the other antibodies.
(3) Remove any antibodies with a concentration below a given threshold and randomly replace with new antibodies from the available pool.
(4) Repeat (2) and (3) until the members of the AIS remain unchangeable for ten iterations.
(5) Generate recommendations based on the final antibody population.

In this model the most important aspect is the concentration, which is calculated as below:

$$\frac{dx_i}{dt} = k_1 m_i x_i y - \frac{k_2}{n}\sum_{j=1}^{n} m_{i,j} x_i x_j - k_3 x_i \tag{1}$$

$k_1$ represents stimulation rate, $k_2$ represents suppression rate and $k_3$ represents death rate
$y$ represents the concentration of the antigen (fixed in our model)
$x_i$ represents the concentration of antibody $i$
$x_j$ represents the concentration of antibody $j$
$m_{i,j}$ represents the affinity between the antibody $i$ and $j$
$m_i$ represents the affinity between the antibody $i$ and the antigen
$n$ represents the number of antibodies in AIS

In our system we fix the parameters as $k_1$=0.3, $k_2$=0.2, $k_3$=0.1, based on Cayzer and Aickelin [3]. For one antibody, the concentration value is proportional to a high antibody-antigen affinity, is decreased with a high antibody-antibody affinity. In the absence of either, the death rate is used to slowly decrease an antibody's concentration. Next we will describe the affinity measure algorithms we use.

## 3  Two Affinity Measure Algorithms

We use two different algorithms, Weighted Kappa (WK) [1] and Kendall's Tau (KT) [12], to calculate the affinity (correlation coefficient). The examples below explain how WK and KT work.

For the examples we select two users from the database. User1 made 50 ratings and User2 made 54 ratings, but we only consider the ratings for the common movies, i.e. those that they both rated. The user encoding includes a movie_id, taken from movie's id in the database, and a rating_score stating how the user feels about the movie. The scale as which the movies are rated is as follows: 0, 0.2, 0.4, 0.6, 0.8, 1; where 0 is very bad, 0.2 is bad, 0.4 is below average, 0.6 is above average, 0.8 is good, and 1 is very good.

The two users' encodings are shown below: Generic: {(movie_id, rating_score)}

User1: {(153, 0.6), (253, 0.6), (296, 1), (349, 0.8), (355, 0.4), (457, 1), (553,1), (595,1)}

User2: {(153, 0.8), (253, 0.8), (296, 0.4), (349, 0.8), (355, 0), (457, 0.8), (553, 0.6), (595, 0.8)}

## 3.1 Weighted Kappa (WK)

Before we use the WK Algorithm to calculate the affinity of the two users mentioned above, we note that User1 and User2 have eight movies in common, thus the observation number is 8. We use Table 1 below to summarise the user information.

Table 1. Frequencies ($f_{ij}$) Table for Example 1:

| User1 \ User2 | Very Bad (1) | Bad (2) | Below Average (3) | Above Average (4) | Good (5) | Very Good (7) | Total |
|---|---|---|---|---|---|---|---|
| Very Bad (1) | 0 ($f_{11}$) | 0 ($f_{12}$) | 0 ($f_{13}$) | 0 ($f_{14}$) | 0 ($f_{15}$) | 0 ($f_{16}$) | 0 |
| Bad (2) | 0 ($f_{21}$) | 0 ($f_{22}$) | 0 ($f_{23}$) | 0 ($f_{24}$) | 0 ($f_{25}$) | 0 ($f_{26}$) | 0 |
| Below Average (3) | 1 ($f_{31}$) | 0 ($f_{32}$) | 0 ($f_{33}$) | 0 ($f_{34}$) | 0 ($f_{35}$) | 0 ($f_{36}$) | 1 |
| Above Average (4) | 0 ($f_{41}$) | 0 ($f_{42}$) | 0 ($f_{43}$) | 0 ($f_{44}$) | 2 ($f_{45}$) | 0 ($f_{46}$) | 2 |
| Good (5) | 0 ($f_{51}$) | 0 ($f_{52}$) | 0 ($f_{53}$) | 0 ($f_{54}$) | 1 ($f_{55}$) | 0 ($f_{56}$) | 1 |
| Very Good (6) | 0 ($f_{61}$) | 0 ($f_{62}$) | 1 ($f_{63}$) | 1 ($f_{64}$) | 2 ($f_{65}$) | 0 ($f_{66}$) | 4 |
| Total | 1 | 0 | 1 | 1 | 5 | 0 | 8 |

$f_{ij}$ (frequency in row $i$ and column $j$ in Table 1) is used to calculate the affinity of the two users. It represents how many movies user1 rated as category $i$, and user2 rated as category $j$, $i,j \in \{1,2,3,4,5,6\}$, where category 1= 'Very Bad', category 2= 'Bad', category 3='Below Average', category 4= 'Above Average', category 5= 'Good', category 6= 'Very Good'.

We use Equation (2), which represents the Weighted Kappa Algorithm, to calculate the affinity (correlation coefficient) from the observed and expected frequencies as:

$$WK = \frac{p_{o(w)} - p_{e(w)}}{1 - p_{e(w)}} \quad (2)$$

where *WK* represents the Weighted Kappa value (affinity); $P_{o(w)}$ represents the observed frequencies and $P_{e(w)}$ represents the expected frequencies obtained by chance.

In our system, all users choose the movies they have seen from the database and rate them. Hence, no frequencies exist by chance, and thus $P_{e(w)} = 0$ and $WK = P_{o(w)}$. $P_{o(w)}$ can be calculated by Equation (3).

$$p_{o(w)} = \frac{1}{n} \sum_{i=1}^{g} \sum_{j=1}^{g} w_{ij} f_{ij} \quad (3)$$

where $P_{o(w)}$ represents the observed frequencies; $g$ represents the number of categories; $n$ represents the number of observations in $g$ categories; $f_{ij}$ represents the number of frequencies for the cell (of Table 1) in row $i$, column $j$ and $w_{ij}$ represents the weight value for the cell (of Table 1) in row $i$ and column $j$.

Here, the number of categories is six, $g = 6$; the observation number $n$ is the number of movies that two users have in common; $f_{ij}$ can be acquired from Table 1.

Finally, $w_{ij}$ can be obtained from Equation (4) [1]. The smaller the difference between $i$ and $j$, the greater the weight as the two users have a stronger agreement on this movie and vice versa. When $i=j$, the weight will reach the greatest value of 1 as the two users have the strongest agreement:

$$w_{ij} = 1 - \frac{|i - j|}{g - 1} \quad (4)$$

where $w_{ij}$ is the weight value for the cell of Table 1 in row $i$ and column $j$; hence $i$ and $j$ also respectively represent the rating category of User1 and User2 for the same movie. Using the above, we can calculate WK for User1 and User2 as:

$$WK(user1, user2) = \frac{1}{8} \times (0.6 \times 1 + 0.8 \times 2 + 1 \times 1 + 0.4 \times 1 + 0.6 \times 1 + 0.8 \times 2) = 0.725$$

The WK correlation between User1 and User2 is: $k_{(w)} = P_{o(w)} = 0.725$ (i.e. good agreement).

### 3.2 Kendall's Tau method (KT)

Central to KT is the notion of concordance and discordance. We view pairs of observations $(X_i, Y_i)$ and $(X_j, Y_j)$ as concordant if $X_j-X_i$ and $Y_j-Y_i$ have the same sign and as a discordant if $X_j-X_i$ and $Y_j-Y_i$ have opposite signs. Let $C$ represent the number of concordant and $D$ represent the number of discordant pairs. KT is then as defined in Equation (5).

$$KT = \frac{2S}{n(n-1)}, -1 \leq KT \leq +1 \tag{5}$$

Where $n$ represents observation number; $S$ represents Kendall's S, defined as $S = C-D$. For $n$ observations, there are $n(n-1)/2$ pairs ($1 \leq i \leq j \leq n$). If they are all concordant, $KT$ is 1, if they are all discordant $KT$ is -1. $C$ is the number of concordant pairs; $D$ is the number of discordant pairs.

Using the example of User1 and User2, the table below shows all the concordant pairs and discordant pairs of observations. We view zero and zero as a concordant pair. Zero and other non-zero numbers as neither discordant nor concordant and we ignore them. Otherwise, the KT will differ depending on the order the values arrive in. This would produce a distorted result.

Table 2. Sample calculations of Kendall's Tau

| (Movie-id, Movie-id) | User1 | User2 | Decision |
|---|---|---|---|
| (153, 253) | 0.6-0.6 = 0 | 0.8-0.8 = 0 | Concordant |
| (153, 296) | 0.6-1 = -0.4 | 0.8-0.4 = 0.4 | Discordant |
| (153, 349) | 0.6-0.8 = -0.2 | 0.8-0.8 = 0 | n/a |
| (153, 355) | 0.6-0.4 = 0.2 | 0.8-0 = 0.8 | Concordant |
| (153, 457) | 0.6-1 = -0.4 | 0.8-0.8 = 0 | n/a |
| (153, 553) | 0.6-1 = -0.4 | 0.8-0.6 = 0.2 | Discordant |
| (153, 595) | 0.6-1 = -0.4 | 0.8-0.8 = 0 | n/a |

If we calculate all values, we will find that there are 9 concordant pairs, 6 discordant pairs, thus C = 9, D = 6; S =9-6=3; KT = (2*3)/8*(8-1) =0.1071 (i.e. some agreement).

## 4 System Implementation

Following the algorithmic outline given in section 2, we construct two movie recommenders: One uses WK, the other KT as the affinity algorithm of the AIS. After the AIS has chosen 100 users (antibodies), who have similar preferences to the user (antigen) who requires recommendations, the CF algorithm uses Equation (6) below to calculate the predictions:

$$prediction = \frac{\sum_{i=1}^{100}(weight_i \times Rating_i)}{\sum_{i=1}^{100}(weight_i \times 1)} \tag{6}$$

Where $weight_i$ represents the concentration of the $ith$ antibody and $Rating_i$ represents the rating, which the $ith$ antibody gave the movie. We use the concentration as the weight to calculate the prediction, as it contains both the correlation of the antibody to the antigen and the correlation of the antibody to the other antibodies.

The data used in this project is publicly available data, which is offered by Compaq Research (formerly DEC Research) [6]. It contains 2,811,983 ratings entered by 72,916 users for 1,628 different movies, and it has been used in numerous CF publications. We randomly chose 4,000 users whose user Ids are larger than 15,000 and who have rated more than 20 movies as candidate antibodies. We use the users whose Person Ids are smaller than 15,000 as the test target users or antigens.

## 5  Experimental Results

All experiments were executed on an Intel Pentium 4 CPU, 1.5G Hz, 256MB RAM, Windows 2000 platform computer. The system was coded using Java 2 platform, Standard Edition (J2SE) 1.4.0. The database was implemented using Microsoft Access (XP Professional). Using the above experimental environment, the execution time for one user to obtain recommendations is several seconds.

### 5.1 Experiment 1 – Ties in KT

The objective of this experiment was to find out how much information we ignore when we use KT, because of the ties, and whether this seems reasonable. In this experiment we calculated the ignored percentage of votes of 350 users, randomly chosen from the database. Figure 1 shows that on average 38.27% of the information was ignored. On some occasions, more than 50% of the information was ignored. On average, this represents a large amount of data and results might therefore be adversely affected when KT is used as an affinity measure.

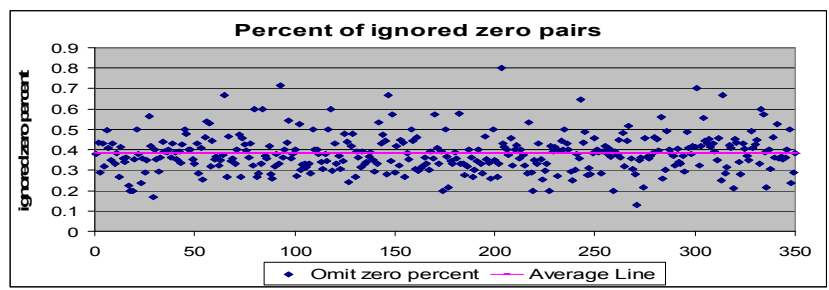

Figure. 1. Percentage of ignored votes (KT).

### 5.2 Experiment 2 – Recommendation Accuracy of KT versus WK

The objective of this experiment was to ascertain whether the choice of affinity measure affects the results. We calculated the prediction accuracy for 500 users (who rated more than 20 movies). For each user we hid one rating, and made the prediction for the hidden movie using the remaining information. We repeated this 20 times for each user and compare the 20 predictions with their hidden actual ratings. We obtained the prediction accuracy for each user as:

Prediction accuracy = 1 – (|prediction – actual rating| / 20).

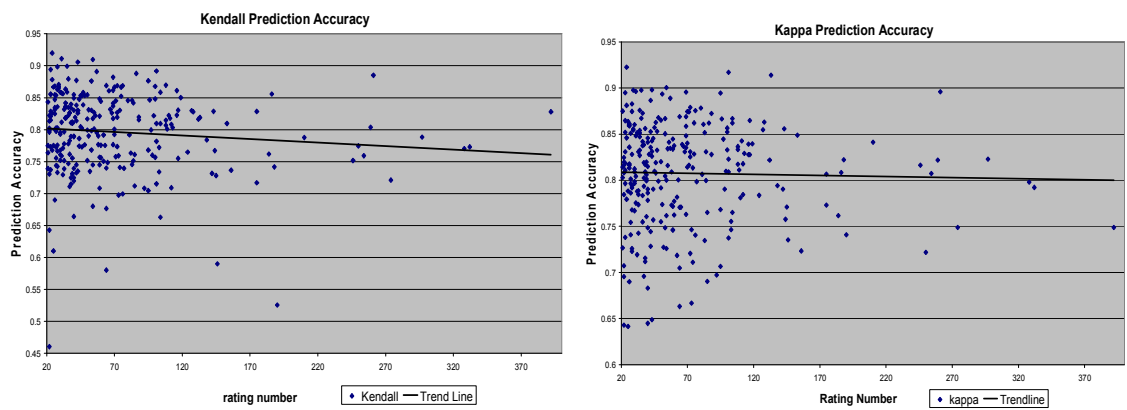

Figure. 2. Prediction accuracy for KT versus prediction accuracy for WK.

The left-hand graph in Figure 2 shows the prediction accuracy for 500 users using KT correlation. The median prediction accuracy was 0.798. The trend line shows that the prediction accuracy of users decreases slightly as the rating numbers increase.

The right-hand graph in Figure 2 shows the prediction accuracy for 500 users using the WK correlation method. The trend line shows that rating numbers do not affect the users' prediction accuracy. The median prediction accuracy was 0.802. To put this in layman's terms, an accuracy of 0.802 indicates that there is one rank difference between the prediction and the user's actual rating. For example if the system predicts a movie as 'Very Good', the user may think that it is 'Good'. Although WK performed slightly better, using a t-test to analyze the data shown in Figure 2, we found that there is no significant difference between the prediction accuracy obtained from the two methods.

As we use the same parameters and algorithmic framework as in previous work [3], we can accurately compare our results to those found there. In our experiments, the median prediction accuracy for WK was 0.802 and 0.798 for KT. In the previous work, the affinity measure used is either a simple correlation method (SC) or Pearson's Rank Correlation (PR). The median prediction accuracy of those was reported as 0.652 (SC) and 0.697 (PR). Hence, we can rightly claim that by using a more appropriate affinity function, we significantly improve upon previous results.

## 6  Discussion and Conclusions

In this paper we have shown that using a more appropriate affinity measure, such as KT or WK, we can significantly improve upon movie recommendation results reported earlier. Furthermore, the AIS has shown itself to be relatively robust to the affinity algorithm chosen. Looking more closely at our KT findings, we ignore ties, which leads to a loss of information (on average 38.27%). This is caused by the fact that in the movie prediction problem there are only six categories, but more possible comparisons (for *n* categories there will be ½*n(n-1) comparisons).

For this reason, if we use KT, the recommendation system is likely to obtain worse results than for WK. However, the test results show that the prediction accuracy using these two methods has only a slight difference, with KT being only marginally worse than WK. We assume this is because there is a large amount of user information in our system (70,000 users) and the algorithm can continue until enough suitable users are identified. Thus ignoring large amounts of data is less of a problem, but leads to longer run-times. However, if there is less user information or if run-time was crucial, significantly better results could almost certainly be obtained using WK rather than KT. Therefore, we can conclude that in general WK is better than KT as an affinity measure in movie recommendation.